%% file: emnlp2021.tex
\pdfoutput=1
\documentclass[11pt]{article}

\usepackage{emnlp2021}

\usepackage{times}
\usepackage{latexsym}
\input{formatting/math_commands}

\usepackage[T1]{fontenc}
\usepackage{ragged2e}
\usepackage{siunitx}
\usepackage{adjustbox}
\newcommand*\rot{\rotatebox{90}}
\newcommand{\STAB}[1]{\begin{tabular}{@{}c@{}}\rot{#1}\end{tabular}}

\usepackage{amsfonts}

\usepackage[utf8]{inputenc}

\usepackage{microtype}

\usepackage[subtle]{savetrees}
\usepackage{multirow}
\usepackage{hyperref}
\usepackage{booktabs} 
\usepackage{tabularx}
\usepackage{graphicx}
\usepackage{subcaption}

\title{Style Pooling: Automatic Text Style Obfuscation for Improved\\ Classification Fairness \vspace{-1ex}}

\author{Fatemehsadat Mireshghallah, \textbf{Taylor Berg-Kirkpatrick}\\
     University of California San Diego, \\
    \texttt{\{fatemeh, tberg\}@ucsd.edu}\\ }

\begin{document}
\maketitle

\setlength{\belowdisplayskip}{1ex} \setlength{\belowdisplayshortskip}{1ex}
\setlength{\abovedisplayskip}{-0.5ex} \setlength{\abovedisplayshortskip}{0pt}

\input{body/abstract}
\input{body/Intro}
\input{body/proposed}

\input{body/exp-setup}

\input{body/exp-results}
\input{body/related}
\input{body/conclusion}
\input{body/ackn}
\bibliography{anthology,custom}
\bibliographystyle{acl_natbib}

\clearpage
\appendix

\input{body/appendix}

\end{document}

%% file: formatting/math_commands.tex
\usepackage{amsmath,amsfonts,bm}

\def\eqref#1{equation~\ref{#1}}

\def\1{\bm{1}}

\DeclareMathAlphabet{\mathsfit}{\encodingdefault}{\sfdefault}{m}{sl}
\SetMathAlphabet{\mathsfit}{bold}{\encodingdefault}{\sfdefault}{bx}{n}

\def\gL{{\mathcal{L}}}

\def\sE{{\mathbb{E}}}

\newcommand{\KL}{D_{\mathrm{KL}}}

\newcommand{\xc}{X}

\newcommand{\x}{x}

%% file: body/abstract.tex
\vspace{-1.5ex}
\begin{abstract}
\vspace{-1.5ex}
 Text style can reveal sensitive attributes of the author (e.g. race or age) to the reader, which can, in turn, lead to privacy violations and bias in both human and algorithmic decisions based on text. For example, the style of writing in job applications might reveal protected attributes of the candidate which could lead to bias in hiring decisions, regardless of whether hiring decisions are made algorithmically or by humans. We propose a VAE-based framework that obfuscates stylistic features of human-generated text through style transfer \textit{by automatically re-writing the text itself}. Our framework operationalizes the notion of obfuscated style in a flexible way that enables two distinct notions of obfuscated style: (1) a minimal notion that effectively \textit{intersects} the various styles seen in training, and (2) a maximal notion that seeks to obfuscate by adding stylistic features of \textit{all sensitive attributes} to text, in effect, computing a \textit{union} of styles. Our style-obfuscation framework can be used for multiple purposes, however, we demonstrate its effectiveness in improving the fairness of downstream classifiers. We also conduct a comprehensive study on style pooling's effect on fluency, semantic consistency, and attribute removal from text, in two and three domain style obfuscation.\footnote{Code, models, and data is available at \url{https://github.com/mireshghallah/style-pooling}}                                        
\end{abstract}

%% file: body/Intro.tex
\vspace{-3.5ex}
\section{Introduction}
\vspace{-1ex}
Machine learning (ML) algorithms are used in a wide range of tasks, including high-stakes applications like determining credit ratings, setting insurance policy rates, making hiring decisions, and performing facial recognition. It has been shown that such algorithms can produce outcomes that are biased towards a certain gender or race~\cite{buolamwini2018gender, silva-etal-2021-towards, sheng2021societal}. 
%

\begin{table*}[]
    \centering
    \caption{Example Blog sentences transformed with A4NT~\cite{a4nt} and our proposed Intersection and Union obfuscations. Our Intersection obfuscation aims at changing the style such that it does not reflect either teen or adult style. However, the union, tries to reflect both by making changes like adding ``...'' to the beginning of the sentence (adult style) while keeping the ``grr'' (teen style). Or by adding exclamation marks at the end of the sentence.}
    \vspace{-1ex}
    \label{tab:sentences}
    \centering
    \begin{adjustbox}{width=1\linewidth, margin=-1ex 0ex 0ex 0ex}
    \input{tables/sentences_all}
    \end{adjustbox}
    \vspace{-3ex}
\end{table*}
%
Ideally, high-stakes decisions made by either humans or ML algorithms, should not be influenced by irrelevant, protected attributes like nationality, age, or gender.
In many instances, the input data used for making high-stakes decisions is text that is authored by a human candidate -- for example, hiring decisions are often based on bios and personal statements. 
Recent work~\cite{de2019bias} shows that automatic hiring-decision models trained on bios are less likely to select female candidates for certain roles (e.g. architect, software engineer, and surgeon) even when the gender of the author is not explicitly provided to the system. Bias is, of course, not limited to algorithmic decisions, humans make biased decisions based on text, even when the protected attributes of the author are not explicitly revealed~\cite{pedreshi2008discrimination}. 
Together, these results indicate that both algorithms and humans can (1) decipher protected attributes of authors based on stylistic features of text, and (2) whether consciously or not, be biased by this information. 

A large body of prior work has attempted to address \textit{algorithmic bias} by modifying different stages of the natural language processing (NLP) pipeline. For example, \citet{ravfogel2020null} attempt to de-bias word embeddings used by NLP systems, while \citet{elazar2018adversarial} address the bias in learned  model representations and encodings. 
While effective in many cases, such approaches do nothing to mitigate bias in decisions made by humans based on text. We propose a fundamentally different approach. Rather than mitigating bias in learning algorithms that make decisions based on text, we propose a framework that  obfuscates stylistic features of human-generated text \textit{by automatically re-writing the text itself}. By obfuscating stylistic features, readers (human or algorithms) will be less able to infer protected attributes that enable bias.

We introduce a novel framework that enables `style pooling': the automatic transduction of user-generated text to a central, obfuscated style. Notions of `centrality' can themselves introduce bias -- for example, a system might learn to obfuscate by mapping all text to the dominant style seen in its training corpus. This might `white-wash' text, ignoring stylistic features of underrepresented groups in the learned notion of central style. Our framework operationalizes the notion of centrality in a more flexible way: our probabilistic approach allows us to choose between two distinct notions of centrality.
First, we define a variant of our model which is incentivized to learn a minimal notion of central style that effectively \textit{intersects} the various styles seen in training. This is achieved through  the design of this variant's probabilistic prior.
We further equip this variant with a novel ``de-boosting'' mechanism, which amplifies the use of words that are less likely to leak sensitive attributes, and de-incentivizes the use of words whose presence might hint at a particular sensitive attribute.
Second, we propose an alternative prior that instead incentivizes a maximal notion of style that seeks to obfuscate by adding stylistic features of all protected attributes to text -- in effect, computing a \textit{union} of styles.
Table~\ref{tab:sentences} shows our intersection and union obfuscation applied to sentences from the Blogs dataset, and highlights the differences between them.
%

While we propose both these obfuscations in our framework and leave it to the users to choose, it is worth noting that the cognitive process literature shows that when humans are confronted with conflicting biasing information, they tend to form an opinion about the conflicting text, based on their own implicit biases~\cite{richter2017comprehension}. Therefore, removing sensitive stylistic features may be more effective than combining them. 
This is also commensurate with our findings, where we observed that intersection  more successfully improves the fairness metric (Section~\ref{sec:fairness}).
%
%

We extensively evaluate our proposed framework on a wide range of tasks. First, we compare and contrast our ``intersection'' and ``union'' obfuscations on a modified version of the Yelp dataset~\cite{shen2017style} where we have created three stylistic domains by deliberately misspelling three disjoint sets of words. We show that our intersection obfuscation successfully removes these misspellings and replaces them by the dominant spelling of the word $99.20\%$ of the time, while our union obfuscation spreads the misspellings into the other two domains $46.40\%$ of the time. 
Then, we evaluate our framework on the Blogs data~\cite{schler2006effects}, where the sensitive attribute is age, and we measure the impact our obfuscations have on the fairness of a job classifier, using the the TPR-gap measure from~\citet{de2019bias}.
%
We also evaluate the removal of sensitive attributes, fluency of the generated text, and the uncertainty of a sensitive attribute classifier for our framework, in both two and three domain setups. 
%
%

%


%

%% file: tables/sentences_all.tex
 \newcolumntype{L}{>{\centering\arraybackslash}m{0.07\linewidth}} 
  \newcolumntype{O}{>{\centering\arraybackslash}m{0.08\linewidth}} 
  \newcolumntype{D}{>{\arraybackslash}m{0.20\linewidth}} 
  \newcolumntype{R}{>{\arraybackslash}m{0.29\linewidth}} 
  \centering
\begin{tabular}{@{}lllll@{}}
	\toprule
    Age &Input Sentence (Original Data)&A4NT (Baseline)& Intersection &Union \\
    \midrule
Teen&\textbf{grr} ... now i get cold quicker . & \textbf{grr} now i get cold \textbf{lol} . & \textbf{hmmm} ... now i get cold .  & \textbf{... grr} ... now i get cold quicker .   \\
Teen&it was so \textbf{fricken} hilarious .     & it was so ~\textbf{boring} hilarious .   &  it was so ~\textbf{utterly} hilarious . & it was so ~\textbf{ totally} hilarious  \\
Adult&well i 've just been ~\textbf{too busy} .                  &well i 've just been ~\textbf{kinda fun} . & well i 've just been ~\textbf{too busy} .   &well i 've just been ~\textbf{too busy .} \\
Adult& these were ~\textbf{common phrases} .    &  these were ~\textbf{common teacher} .  &   these were~\textbf{ common} .  &these were ~\textbf{common ! !}\\
	\bottomrule
\end{tabular}

%% file: body/proposed.tex
\vspace{-1ex}
\section{Proposed Method}
\vspace{-1.5ex}

\begin{figure}[!t]
    \includegraphics[width=\linewidth]{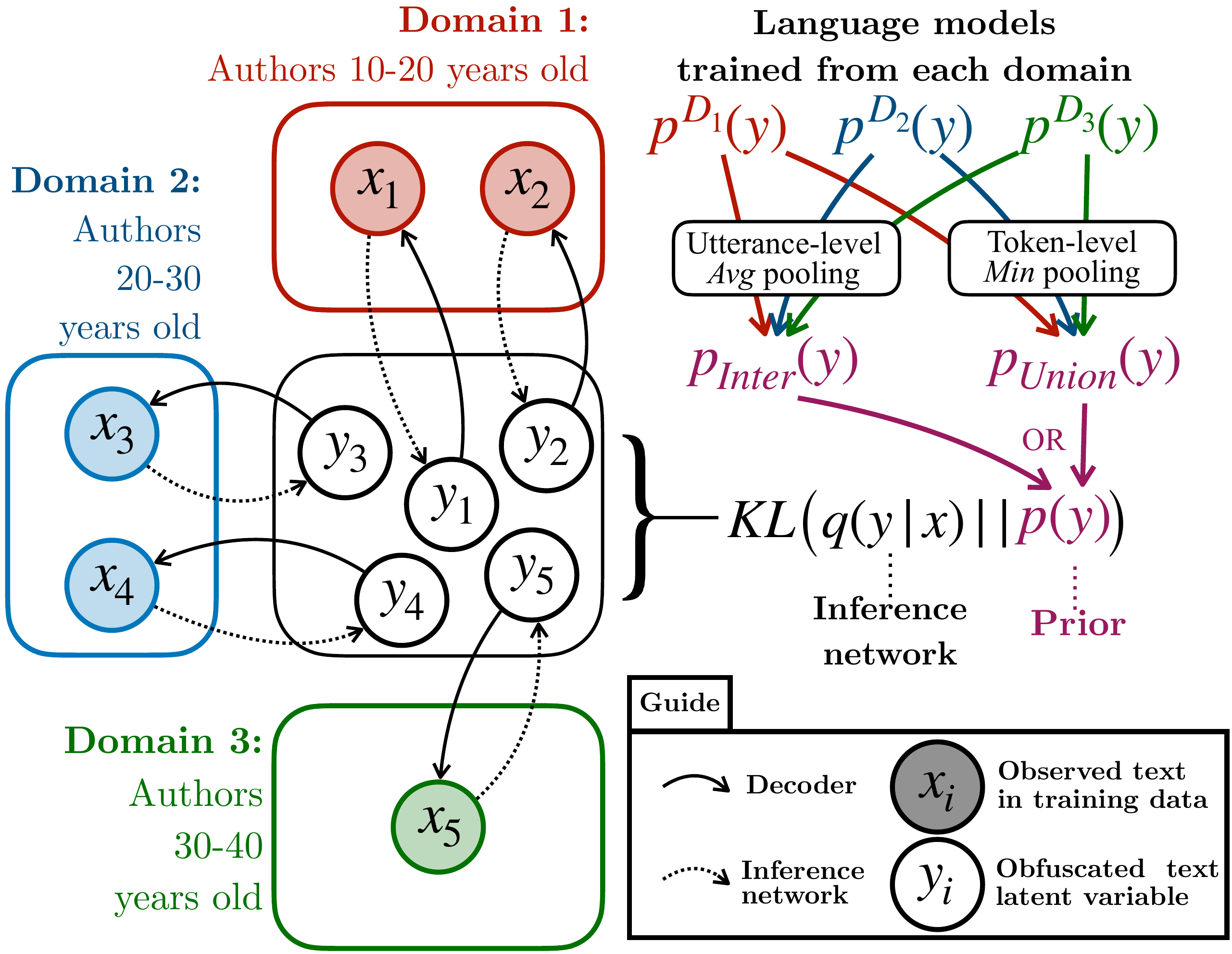}
    \caption{Proposed unsupervised framework for  \textit{style pooling}: inducing a centralized obfuscated style. $x_i$ represent observed text which are clustered by  their sensitive attribute (age). $y_i$ are corresponding latent variables representing the induced obfuscated text. Training leverages an amortized inference setup similar to a VAE-style training, but, critically the prior is produced by pooling language models from each domain using two different strategies targeting (1) intersected style and (2) the union of all styles in the corpus. }
    \vspace{-3ex}
    \label{fig:model}
\end{figure}

 In this section, we  first introduce our model structure, then describe our style-pooling priors and the unsupervised learning and inference techniques we leverage for this model class. Finally, we introduce our style de-boosting mechanism. 
\vspace{-1ex}
\subsection{Model Structure}
\vspace{-1ex}
Consider a training corpus consisting of utterances produced by authors with various protected attributes. In Figure~\ref{fig:model}, we depict a grouping of authors by age into three domains. We let $x_i$ represent an individual observed text utterance in the corpus, and assume $M$ domains (sensitive attribute classes) in the dataset.
$y_i$ is a latent variable that represents the obfuscated version of $x_i$. Hence, $y_i$ is a text valued latent, while $x_i$ is a text valued observation.
%
%
We let $d(i)$ denote the domain of the $i^{th}$ sample in the dataset. 
%
With this definition, our generative process assumes each sentence $x_i$, with corresponding domain $d(i)$, is generated as follows: First, a latent sentence $y_i$ is sampled from a central prior, $p_{prior}(y_i)$, which is domain agnostic. Then, $x_i$ is sampled conditioned on $y_i$ from a transduction model, $p(x_i|y_i;\theta_{y \to x}^{d(i)})$ . We let $\theta_{y \to x}^{D_j}$ represent the parameters of the transduction model for the $j$th domain. We extensively discuss $p_{prior}$ in the next section. For now, we assume the prior distributions are pretrained on the observed data and therefore omit their parameters for simplicity of notation. Together, this gives the following joint likelihood:

{\small
\vspace{-1ex}
\begin{equation}
\begin{alignedat}{1}
\label{eq:joint}
p(&\xc^{D_1},...,\xc^{D_M}, Y;\theta_{y \to x}^{D_1}, ...,\theta_{y \to x}^{D_M})\\ &=  \prod^N_{i = 1}p\big(\x_i|y_{i}; \theta_{y \to x}^{d(i)}) p_{prior}\big(y_{i}\big) 
\end{alignedat}
\end{equation}}
The log marginal likelihood of the observed data, which we approximate during training, can be written as:

{\small
\vspace{-1ex}
\begin{equation}
\begin{alignedat}{1}
\label{eq:marginal}
\log\ p(&\xc^{D_1},...,\xc^{D_M};\theta_{y \to x}^{D_1}, ...,\theta_{y \to x}^{D_M})\\= & \log \sum\nolimits_Y p(\xc^{D_1},...,\xc^{D_M};\theta_{y \to x}^{D_1}, ...,\theta_{y \to x}^{D_M})
\end{alignedat}
\end{equation}
}
%
%
\noindent\textbf{Neural Architectures.} We select a parameterization for our transduction distributions that makes no independence assumptions. We use an encoder-decoder architecture based on the standard attentional Seq2Seq model which has been shown to be successful across various tasks~\cite{bahdanau2014neural, rush2015neural}. Our prior distributions for each domain are built using  recurrent language models which also make no independence assumptions.


%
\vspace{-1.5ex}
\subsection{Prior Distributions}
\vspace{-1.ex}
The critical component of our framework that incentivizes obfuscation are our specialized priors, as depicted in Figure~\ref{fig:model}.  We introduce two prior  variants, $p_{\textrm{Inter}}(y)$ and $p_{\textrm{Union}}(y)$, which incentivize induction of intersected styles and the union of all styles, respectively. Each prior is assembled out of $M$ (here $M=3)$ separate language models -- $p^{D_1}$, $p^{D_2}$, ..., $p^{D_M}$ -- each trained on the corresponding domain of observed utterances in the training data. The intersection prior, $p_{\textrm{Inter}}(y)$, is computed by taking the sum of the likelihoods of an entire utterance across the language models from all $M$ domains (and then re-normalizing to ensure that resulting prior is a valid distribution). This utterance-level average pooling approach incentivizes a ``majority-voting'' effect, in which the model is pressured to remove any words and stylistic features that are characteristic of one domain, but not the others, and converge to features that are shared by the majority of the domains. Therefore the prior for intersection becomes:
%

{\small
\vspace{-2ex}
\begin{equation}
    p_{Inter}(y_i) =\frac{1}{M} \sum^M\nolimits_j p^{D_j}(y_i)
    \label{eq:inters}
\end{equation}
}
In contrast, the union prior, $p_{\textrm{Union}}(y)$, computes the likelihood of an utterance according to the minimum likelihood across each domain's language model \textit{at each token position}, $t$.\footnote{The token-wise min of the language models is not, itself, a normalized distribution. However, we can treat it as implicitly normalized in our training objective (discussed in the next section) because the absence of normalization only contributes an additive constant to our objective.} 
Through experimentation (Sec.~\ref{sec:synth}) we empirically observed that this prior rewards the model for inserting as many stylistic features as possible that are unique to each domain.
%

{\begingroup\makeatletter\def\f@size{8}\check@mathfonts
\def\maketag@@@#1{\hbox{\m@th\large\normalfont#1}}%
\vspace{-2ex}
\begin{equation}
\begin{alignedat}{1}
     p_{Union}(y_i) \propto  \prod\nolimits_t^T
     \min (p^{D_1}(y_{i,t}|y_{i,<t}), ..., p^{D_M}(y_{i,t}|y_{i,<t})   )
\end{alignedat}
\vspace{-5ex}
 \label{eq:union}
\end{equation}
\endgroup
}
\vspace{-1ex}
\subsection{Learning and Inference}
\vspace{-1ex}
Training is accomplished using an approach from~\cite{he2020probabilistic}: We employ seq2seq inference networks and use an amortized inference scheme similar to that used in a conventional VAE, but for sequential discrete latents. 

Ideally, learning should directly optimize the log data likelihood, which is the marginal shown in Eq.~\ref{eq:marginal}. However, due to our model's neural parameterization, the marginal is intractable. To overcome the intractability of computing the true data likelihood, we adopt amortized variational inference~\cite{kingma2013auto} to derive a surrogate objective for learning the evidence lower bound (ELBO) on log marginal likelihood:

{\small
\begin{gather}
\begin{alignedat}[b]{2}
  & \log\ p(\xc^{D_1},&&...,\xc^{D_M}; \theta_{y\to x}^{D_1},...,\theta_{y\to x}^{D_M})  \\
 & \geq \gL_{\text{ELBO}} && (\xc^{D_1},...,\xc^{D_M};\theta_{y\to x}^{D_1},...,\theta_{y\to x}^{D_M}, \phi_{x\to y}) \\
 &= \sum^N\nolimits_i  &&\underbrace{\Big[ \sE_{q(y_i | \x_i; \phi_{x \to y})}[\log\ p(\x_i|y_i; \theta_{y \to x}^{d(i)})]}_{\textrm{Reconstruction likelihood}} \\
 & &&- \underbrace{\KL\big(q(y_i | x_i ; \phi_{y \to x})) || p_{prior}(y_i) \big) \Big]}_{\textrm{KL regularizer}} 
\end{alignedat}\label{eq:elbo}
\raisetag{30pt}
\vspace{-2ex}
\end{gather}}

This new objective introduces $q(y | x ; \phi_{x \to y})$, which represents the inference network distribution that approximates the model’s true posterior, $p(y|x; \theta_{x\to y})$. 
Learning operates by optimizing the lower bound over both variational and model parameters. 
Once training is over, the posterior distribution can be used for style obfuscation. 

The reconstruction and KL terms in Eq.~\ref{eq:elbo} involve intractable expectations, which means we need to approximate their gradients. To address this, we use the Gumbel-softmax~\cite{jang2016categorical} straight-through estimator to backpropagate gradients from both the KL and reconstruction loss terms.

\noindent\textbf{Length Control.} During the training of the model, we observed that it tends to repeat the same word when it is trying to generate obfuscated text, $y_i$. To mitigate this, we append two floating point length tokens to the input of the inference networks decoder at each step $t$, one of these tokens tells the model which step it is on, and the other tells it how many steps are left~\cite{kikuchi2016controlling, hu2017toward}. We also experimented with positional embeddings instead of floating point tokens, but we observed that they yield worse convergence. 
Another measure we take to encourage shorter sentences was to hard stop the decoding during training once the re-written sentence had the same length as the original sentence. To further stabilize training we share parameters between the inference network and the transduction models, appending an embedding to the input to indicate the output domain.


%
%
\vspace{-2ex}
\subsection{Style De-boosting}
\vspace{-1ex}
To better encourage the removal of identifying stylistic features, we introduce a de-boosting mechanism, which incentivizes the use of words that are less likely to leak sensitive attributes, and de-incentivizes the use of words whose presence might hint at a particular sensitive attribute. 
%
%
We build on the intuition that for a given word $w$ in the vocabulary, if the probability that it belongs to domain $m$ is similar to the probability that it belongs to domain $k$, for any given $m, k$ within the possible domains, $M$, then we can assume that this word does not reveal style. 
However, if there is a huge gap in the two probabilities, that word might hint at a certain domain if it is present in the re-written text.
%
%
Therefore, we devise a normalized ``style score'', $s$, for each word $w$ in the vocabulary~\footnote{While this style score may also highlight \textit{content} that is characteristic of a domain in addition to stylistic word choices, we find in experiments that our use of de-boosting does not substantially harm the utility of downstream classifiers -- indicating that content is largely preserved, even with de-boosting.}:

\vspace{-1ex}
\begin{equation}
\footnotesize
   s_w = \frac{\max(f^{D_1}_w,f^{D_2}_w, ...,f^{D_M}_w) - \min(f^{D_1}_w,f^{D_2}_w, ...,f^{D_M}_w)}{\max(f^{D_1}_w,f^{D_2}_w, ...,f^{D_M}_w)} \label{word:score}
\end{equation}
Where $f^{D_1}_w$ is frequency of word $w$ in the training corpus for domain $D_1$,  divided  by the overall number of tokens (words) in the domain corpus. 
Using these scores, we modify the output logits of the decoder so that the output probability distribution over the vocabulary for sample $i$ at step $t$ is given by:

\begin{equation}
    p(y_{i,t}|y_{i,<t},x_i) \propto \text{softmax}(L_{i,t} - \gamma*S)
    \label{eq:deboost}
\end{equation}

\noindent Here, $L_{i,t}$ represents the logits at step $t$, while $S$ is the score vector for all the words in the vocabulary. $\gamma$ is a multiplier that helps tune the amount of de-boosting. Due to the nature of this de-boosting mechanism, it makes sense only to use it with the intersection obfuscation and not the union.

%

%% file: body/exp-setup.tex
\section{Experimental Setup}
 Here, we provide a brief description of our experimental setup. Our code, data and model checkpoints are uploaded in the supplementary material. More details on the code, model configurations, datasets and hyperparameters are provided in Appendix Sections~\ref{sec:app-code},~\ref{sec:app-configs},~\ref{sec:app-data} and~\ref{sec:app-hyperparams}.
\subsection{Model Configurations}
We used a single layer attentional LSTM-based Seq2Seq encoder-decoder for all the experiments, with hidden layer size of $512$ for both encoder and decoder, and word embedding size of $128$. For the attribute classifiers and language models, we also use LSTM  models with the same architecture,  with a final projection layer of the size of sensitive classes/vocabulary.
 %
 \subsection{Datasets} \label{sec:exp-data}
 
\noindent\textbf{Synthetic Yelp dataset~\cite{shen2017style}.}   We shuffle all the sentences in the Yelp reviews dataset and divide them into three groups (domains). We then randomly choose $15$ words from the top $20$ highest frequency words in the dataset, and allocate the set of top $5$ words ($W_1$) to $D_1$ (domain $1$), next $5$ to $D_2$ and the least frequent $5$ words to $D_3$. We misspell all occurrences of $W_1$ in $D_1$, by changing ``word'' to ``11word11''. We then add ``11word11'' to the vocabulary, and do this for all the $5$ words in all $3$ domains ($15$ words total).  After this transformation, we have $3$ domains with disjoint stylistic markers, which can help us more concretely analyze our obfuscation mechanism. 
 
  \noindent\textbf{Blogs dataset~\cite{schler2006effects}.}  The blogs dataset is a collection of micro blogs containing over 3.3 million sentences along with annotation of author’s age and occupation. We  use this data in both two and three domain style pooling, where we treat age as the sensitive attribute and balance the data so each domain has the same number of sentences. In the two domain setup,  we divide the data in two groups of teenagers and adults. In the three style setup, we have three groups of teenagers, young adults (20s) and adults (people in their 30s and 40s). 
  We use this dataset for multiple evaluations including fairness. We compare our obfuscation to that of~\citet{a4nt} in all evaluations with this data.

\noindent\textbf{Twitter dataset~\cite{rangel2016overview}.} We use data from the PAN16 dataset, which contains manually annotated (from LinkedIn) age and gender of 436 Twitter users, along with up to 1000 tweets from each user. We use this data for the purpose of sensitive attribute (age) removal comparison with~\citet{elazar2018adversarial} in Section~\ref{sec:app-elazar}, and have therefore used the exact same preprocessing and handling of the data as done by them. 

\noindent\textbf{DIAL dataset~\cite{DIAL}.} This is a Twitter dataset which has binary dialect annotations of African American English (AAE) and Standard American English (SAE)\footnote{Using standard for non-AAE might not be the most suitable naming, but we use it hereon given the lack of a better substitute.}, setting ``author’s race'' as the sensitive attribute. We use this dataset for comparison with the work~\citet{xu2019privacy}.

\subsection{Baselines}
\noindent\textbf{One language model prior (One-LM).}  This model is an instance of our framework which  uses the output distribution of a single language model as the prior. 
%
For the Yelp Synthetic data this single LM is trained on the original data which does not have our modifications and  would provide the ideal ``intersection'', since the original data itself does not have misspellings from any of our synthetic domains and can be considered as central. 
%
In the case of the Blogs data where we don't have any ideal central data which is void of style, we train an age classifier and then choose the sentences from the training set that the classifier missclassifies. We create a new training set with these samples and train a single LM on them, and use it for the prior. The intuition is that if the classifier could not guess the domain, these samples are probably close to the notion of centrality we are looking for.  

\noindent\textbf{A4NT~\cite{a4nt}.} ``A4NT: Author Attribute Anonymity by Adversarial Training of Neural Machine Translation'' is the most closely related past work that also attempts to obfuscate text style through automatic re-writing. However, their adversarial approach uses a discriminator network to hide protected attributes simply by mapping the style of one protected category to that of another. 

\noindent\textbf{PATR~\cite{xu2019privacy}.} Privacy Aware Text Rewriting (PATR) is another work close to ours, which removes sensitive attributes through text re-writing using translation and adversarial learning. 
Unlike style pooling, PATR, targets privacy and is therefore not concerned with the union vs. intersection of sensitive attributes.

\noindent\textbf{Original.} We include an ``original'' baseline in our measurements, which shows the value of a given metric if the original un-obfuscated data was used.  


\subsection{Evaluation Metrics}
Below we discuss our evaluation metrics, all of which are measured on the test data.

\subsubsection{Fairness}
\textbf{TPR-gap.}
%
%
We first define a classifier whose main task is to determine if the occupation of an author is student or not, given text from their blog. We set the age of the author as a sensitive attribute, and want to measure the bias in the classifier, given age. We follow~\cite{de2019bias} and use the ``True Positive Rate gap in age'' metric. This measure quantifies the bias in a classifier by finding the gap between the true positive rate for each sensitive attribute group (teen vs. adult). For a binary sensitive attribute $a$ (age) and a true class (for the classifier's main task) $y$, we define:

\begin{align}
    TPR_{a,y} =& P(\hat{Y} = y | A =  \textsl{a}, Y = y )\\
    GAP_{a,y}^{TPR} =& TPR_{a,y} - TPR_{a',y}
\end{align}

\noindent where $A$ is the random variable denoting binary sensitive attribute with values $a$ and $a'$. $Y$, $\hat{Y}$ are random variables denoting the correct class and the predicted class, respectively. The lower the gap is, the more fair the classifier. We report $GAP_{Teen, Student}$, which reflects how  biased the classifier is towards classifying teens as students.
\subsubsection{Linguistic}
\textbf{Back-Translation (BT) accuracy.} We translate the obfuscated samples back to their original domain using the model, and then for each token see if it has been correctly back-translated to its origin or not. We use this metric to see whether the obfuscated version contains sufficient information about content to reconstruct the original. 

\noindent\textbf{GPT-2 PPL.} We feed our obfuscated test sentences to a huggingface~\cite{gpt-2} pre-trained GPT-2 medium model, and report its perplexity (PPL), as an automatic measure of fluency. Lower PPL hints at more fluent text. 

\noindent\textbf{BLEU Score.} In the Yelp Synthetic data experiments, since we have the original (not misspelled) text, we can calculate and report the BLEU score.

\noindent\textbf{GLEU Score.} We use GLEU~\cite{gleu} score as another metric for evaluating the fluency of the generated sentences.

\noindent\textbf{Lexical Diversity (Lex. Div.)} To better quantify the differences between different obfuscations, we calculate the lexical diversity as a ratio where the  size of the vocabulary of the model's output text is the numerator, and the denominator is the overall size of the model's output text (number of all the tokens in the output). 
%

\subsubsection{Sensitive-Attribute Classification}

\textbf{Sensitive-attribute Classifier (Clsf.) accuracy.} To evaluate the removal of sensitive attributes, we train a sensitive-attribute classifier, and use its accuracy as a metric. The closer the accuracy is to chance level (random guess), the more successful is the removal. 
However, there is a caveat to this metric: it is not always clear how the classifier is making its decision, if it is based on content, or style. Therefore, this metric alone is \textbf{not conclusive}.

\noindent\textbf{Entropy.}
To better measure how uncertain the classifier becomes, we also compute its average Entropy across all test samples. Entropy is always between $[0.0,1.0]$ for two domain classification and $[0.0,1.59]$ for three domain classification. The higher it is, the more uncertain the classifier is (more desirable for our purpose). 

\noindent\textbf{Confident Response (CR) percentage.}
We  calculate the percentage of the responses from the classifier for which it was more than $75\%$ sure.

%% file: body/exp-results.tex
\vspace{-1.0ex}
\section{Experimental Results}
\subsection{Synthetic Yelp Data}\label{sec:synth}

Table~\ref{tab:synthesized} shows the experimental results for the Synthetic Yelp dataset experiment, where we trained our proposed framework using the three synthetic domains with misspellings, as explained in Section~\ref{sec:exp-data}.
The \textit{Corrected}, \textit{Remaining} and \textit{Removed} percentages refer to the average ratio of the misspellings corrected, remaining and removed for each domain. These should all sum up to $100\%$.
The \textit{Spread} is the average ratio of the number of words from one domain that have been changed to misspellings from another domain. For instance, if there are 100 occurrences of ``word'' outside $D_1$ before obfuscation, if $40$ of them are converted to  ``11word11'' after obfuscation, then the spread would be $40\%$. 
The \textit{One-LM}  can be considered as an ``oracle baseline'' in this case, since it was trained on original (no misspellings) data. 

The main goal of this controlled experiment is to compare and contrast our intersection and union obfuscations.
From the Table we can see that both our obfuscations lead to high fidelity (back-translation accuracy) and semantic consistency (BLEU score). They also both render the domain classifier very close to chance level ($33.33\%$).
The main differences between these two methods becomes more clear when we look at the corrected, remaining, and spread numbers. The intersection obfuscation with its average pooling, demonstrates a majority voting behavior which incentivizes correcting the misspellings since 2 out of the 3 language models advocate for the correct spelling. Therefore $99.20\%$ of the misspellings are corrected using intersection, very close to the oracle baseline. 
The Union prior, on the other hand, corrects only $45.17\%$ of the misspellings, and lets $54.37\%$ of them to remain as they are. It also converts $46.40\%$ of the correctly spelled words in other domains to misspellings. This shows that the union is in fact mixing the styles, creating sentences that might have more than one misspelling in them.

\begin{table}
    \centering
    \footnotesize
    \caption{Results for the Synthetic Yelp dataset with $3$ domains. \textit{Corrected} shows what \% of modified words in a domain were corrected back to their original format. \textit{Spread} shows the reverse.}
    \vspace{-2ex}
    \label{tab:synthesized}
    \input{tables/synthesized_multi_dom}
    \vspace{-3ex}
\end{table}

\subsection{Blogs Data}
Tables~\ref{table:ratio},~\ref{tab:2dom} and ~\ref{tab:3dom} summarize the experimental results for the Blogs dataset. Below, we will explain each experiment in more detail. 
\subsubsection{Fairness Results}\label{sec:fairness}
Table~\ref{table:ratio} shows the results for the fairness metric measurements on text generated using different obfuscations, for ``Occupation'' classifiers. We have selected a subset of the Blogs data for this experiment, where author occupation is either student or arts, and the age is either teen or adult (two domain obfuscation). we have taken an approach similar to that of~\citet{ravfogel2020null}, where we create $4$ different levels of imbalance. In all cases, the dataset is  balanced  with respect to both occupation and age. We change only the proportion of each age within each occupation class (e.g., in the 0.8 ratio, the student occupation class is composed of $80\%$ teens and $20\%$ adults, while the arts class is composed of $20\%$ teens and $80\%$ adults). For each imbalance ratio we train the classifier on the original imbalanced data, and then test it with original and autmotically generated data from different baselines.
%

Based on Table~\ref{table:ratio}, we can see that our Intersection obfuscation can improve fairness (TPR-gap) with little harm to the classifier accuracy (Occupation), in comparison to the original data and A4NT. We can trade-off classifier accuracy and fairness, by increasing the de-boosting (DB) multiplier. In the Table, \textit{Intersection} shows Intersection obfuscation with different DB levels. In the case of $DB=40$, we lose slightly more utility, but observe much better fairness. 

A4NT's performance in terms of the fairness metric (TPR-Gap) is comparable to our \textit{Intersection} obfuscation (even without de-boosting), however, in maintaining occupation accuracy (utility), A4NT  performs  much more poorly.  We presume  this is because A4NT  removes sensitive attributes solely based on hints from a discriminator, and the low occupation accuracy suggests   the discriminator captures the content more than it captures style, therefore it changes the meaning and structure of the sentences as well. Our human judgments for semantic consistency and fluency in Section~\ref{sec:human} support this hypothesis. 
%
Our \textit{Union} obfuscation, however, does not improve the fairness. We hypothesize this could be caused by keeping/adding biasing words, which can perpetuate the existing impartialities in the classifier, similar to how human cognition works~\cite{richter2017comprehension}.

\begin{table*}
\caption{Fairness results for the Blogs data. The main task is classifying if the author occupation is student or not. Higher occupation accuracy and lower TPR-gap are better. DB denotes our style de-boosting technique, and the number next to it shows its multiplier. Larger multiplier means stronger style obfuscation. }\label{table:ratio}
    \vspace{-2ex}
   \centering
   \footnotesize
   \begin{adjustbox}{width=\textwidth, center}
\input{tables/fairness_ratio_comp_temp}
    \end{adjustbox}
    \vspace{-1ex}
\end{table*}
\subsubsection{Linguistic and Sensitive-attribute Classification Results}
The top section of Tables~\ref{tab:2dom} and~\ref{tab:3dom} show the linguistic and sensitive-attribute classification metrics for the two and three domain obfuscations, respectively. Since A4NT cannot be applied to non-binary style obfuscations as is, there are no results for it in three domains. 
We can see that for both two and three domains the de-boosting (denoted as DB) offers a trade-off between the linguistic quality of the generated text and the obfuscation of sensitive attributes.
Compared to the One-LM baseline, for corresponding levels of de-boosting, our \textit{Intersection} obfuscation is almost always superior, in both text quality and obfuscation. 
%
%
The \textit{Intersection} obfuscation with de-boosting multiplier of 25 outperforms A4NT, with lower classifier accuracy, higher entropy and much lower Confident Response (CR) rate from the classifier. In general, the \textit{Intersection} obfuscation, even without de-boosting does well on \textit{Entropy} and \textit{CR}, which shows that our method is doing well at creating doubt in terms of what the age of the author is.
One caveat however, across both two and three domain obfuscations is the classifier accuracy, which does not decrease much. We hypothesize that one reason for this could be the dependency between style and content, and that the sensitive-attribute classifier could be basing its decisions on content, therefore changing the style would not hide the sensitive attribute.  
Our \textit{Union} obfuscation is behaving differently from the \textit{Intersection}, and is inferior in terms of obfuscating the text, with higher classifier accuracy and lower entropy. However, it has higher lexical diversity, which could hint at it trying to keep sentences diverse and ``adding styles", whereas the \textit{Intersection} is only keeping the common words and is therefore decreasing the lexical diversity. 



%
\begin{table}[]
    \centering
    \caption{Comparison with PATR~\cite{xu2019privacy}, on the Twitter DIAL dataset, where the author's race is the sensitive attribute.}
    \vspace{-1ex}
    \label{tab:patr}
    \begin{adjustbox}{width=\linewidth, center}
    \input{tables/PATR}
    \end{adjustbox}
    \vspace{-1ex}
\end{table}
\begin{table*}[]
    \centering
    \caption{Linguistic and sensitive-attribute classifier results for Blogs data, considering \textit{two} sensitive age domains of teens and adults. For BT accuracy and entropy  higher is better, for PPL and  Confident Response (CR) lower is better.}
    \vspace{-2ex}
    \label{tab:2dom}
    \begin{adjustbox}{width=\textwidth, center}
    \input{tables/2dom}
    \end{adjustbox}
\end{table*}
\begin{table*}[]
    \centering
    \caption{Linguistic and sensitive-attribute classifier results for Blogs data, considering \textit{three} sensitive age domains of teens and adults. For BT accuracy and entropy  higher is better, for PPL and  Confident Response (CR) lower is better. }
    \vspace{-2ex}
    \label{tab:3dom}
    \begin{adjustbox}{width=\textwidth, center}
    \input{tables/3dom}
    \end{adjustbox}
    \vspace{-2ex}
\end{table*}

\subsection{Comparison with PATR}
Table~\ref{tab:patr} provides a comparison between our style pooling method, and PATR~\cite{xu2019privacy}.
$\alpha$ is knob used by PATR to tune the intensity of attribute removal, and the classifier accuracy on non-modified text is 86.3\%.
We can see that without de-boosting, our intersection method drops the classifier accuracy to 74.05\% with a GLEU score of 26.32. PATR drops the classifier accuracy to 74.85\%, but with a worse level of GLEU. With de-boosting, however, we can achieve a classifier accuracy of 62.12\% with GLEU of 17.2, whereas PATR reports accuracy of 65.75\% for a much lower GLEU of 9.67 when $\alpha$ is increased. This shows that our de-boosting mechanism can provide an advantage by giving a lower probability to attribute revealing components, while maintaining the sentence structure. Our union method also achieves 73.27\% accuracy with 26.25 GLEU, making it most suitable for cases where the semantic consistency of the sentences is most important.

\subsection{Evaluation with Human Judgments}
\label{sec:human}

We design two crowd-sourcing tasks on Amazon Mechanical Turk. (1) Fluency: We provide workers with a pair of obfuscated sentences, and ask them which sentence is more fluent. (2) Semantic Consistency: We provide the original (un-obfuscated) sentences, and ask workers which of the obfuscated sentences is closer in meaning to the original sentence.
The model checkpoints used for human evaluations here are those whose fairness and linguistic metrics are reported in Tables~\ref{table:ratio},~\ref{tab:2dom}. We use our intersection obfuscation, with no de-boosting. We  randomly select $188$ sentences from the test set, and used the model outputs for human judgment.
For consistency, each pair of sentences is rated by three workers and we take the majority vote.
In terms of fluency, the workers preferred our obfuscations over those of A4NT for  $60.38\%$ of the sentences. In terms of semantic consistency, for $72.13\%$ sentences they found our obfuscations to be closer in meaning to the original ones. 

%


%% file: tables/synthesized_multi_dom.tex
 \newcolumntype{L}{>{\centering\arraybackslash}m{0.07\linewidth}} 
  \newcolumntype{O}{>{\centering\arraybackslash}m{0.08\linewidth}} 
  \newcolumntype{D}{>{\centering\arraybackslash}m{0.18\linewidth}} 
  \newcolumntype{R}{>{\arraybackslash}m{0.29\linewidth}} 
\begin{tabular}{@{}lSSS@{}}
	\toprule
		&{Intersection}&	{Union}&	{One-LM} \\
    \midrule
	BT Accuracy (\%)&	92.47&	94.52&	95.58	\\
    Corrected (\%)&	99.20&	45.17&	99.87\\
    Remaining (\%)&	0.61&	54.37&	0.00 \\
    Removed (\%)&	0.18&	0.46&	0.12\\
    Spread (\%)&	0.18&	46.40&	0.00\\
    Cls Accuracy (\%)&	33.48&	34.99&	33.35\\
    BLEU	&81.74&	70.86	&93.01\\
	\bottomrule
\end{tabular}

%% file: tables/fairness_ratio_comp_temp.tex
\centering
\begin{tabular}{@{}lSSSSSScSSSSSS@{}}
\toprule
\multirow{4}{*}{Ratio} & \multicolumn{6}{c}{Occupation Accuracy (Utility)} & & \multicolumn{6}{c}{TPR-gap (Fairness)}  \\\cmidrule{2-7} \cmidrule{9-14} 
                        & {\multirow{2}{*}{Original}}     & {\multirow{2}{*}{A4NT}} &\multicolumn{3}{c}{\text{Intersection}} &{\multirow{2}{*}{Union}} &   & {\multirow{2}{*}{Original}}     & {\multirow{2}{*}{A4NT}} &\multicolumn{3}{c}{\text{Intersection}}  &{\multirow{2}{*}{Union}}\\ \cmidrule{4-6} \cmidrule{11-13}
                        &              &          & \text{No DB} &   \text{DB= 25}       & \text{DB= 40}     &       &              &     &&    \text{No DB} &   \text{DB= 25}       & \text{DB= 40}       &        \\\midrule
$0.95$                  & 74.56        & 59.35    & 74.77 &  71.12 &69.73       &73.22 &       & 0.54         & 0.29    &0.23  &   0.23    & 0.21     &0.51\\
$0.80$                  &65.55	       &54.74	  & 65.31 &  65.12 &59.60       &65.43 &       &0.35	        &0.21	  &0.35 &    0.18    &0.18      & 0.36\\
$0.65$                  &59.01	       &52.73	  & 58.41 &  56.68 &54.45       &57.19 &       &0.12	        &0.05	 &0.11  &   0.11    &0.11       &0.15\\
$0.50$                  &58.09	       &53.47	  & 56.21 &  53.6 &53.18	  &55.49 &       &0.04	       &0.08	&0.05 &     0.05    &0.03        &0.05\\\bottomrule

                          
\end{tabular}

%% file: tables/PATR.tex
 \newcolumntype{L}{>{\RaggedLeft\arraybackslash}p{0.06\linewidth}} 
  \newcolumntype{O}{>{\RaggedLeft\arraybackslash}m{0.07\linewidth}} 
  \newcolumntype{D}{>{\arraybackslash}m{0.15\linewidth}} 
  \newcolumntype{R}{>{\arraybackslash}m{0.29\linewidth}} 
\begin{tabular}{@{}llSSSSScSSSS@{}}
	\toprule
	 {\multirow{2}{*}{Metric}}  &\multicolumn{2}{c}{PATR}	&   {}& \multicolumn{2}{c}{Intersection}	& {\multirow{2}{*}{Union}} \\
	\cmidrule{2-3} \cmidrule{5-6} 
	&                                  {$\alpha=1$} &  {$\alpha=5$}          &                   & {No DB} &  {DB = 20}  \\
    \midrule
    GLEU    & 24.77 &9.67&	&26.32	&17.21	&26.25		 \\
    Clsf. Acc (\%)	&74.85&65.75&	&74.05	&62.12	&73.27		\\
	\bottomrule
\end{tabular}

%% file: tables/2dom.tex
 \newcolumntype{L}{>{\RaggedLeft\arraybackslash}p{0.06\linewidth}} 
  \newcolumntype{O}{>{\RaggedLeft\arraybackslash}m{0.07\linewidth}} 
  \newcolumntype{D}{>{\arraybackslash}m{0.15\linewidth}} 
  \newcolumntype{R}{>{\arraybackslash}m{0.29\linewidth}} 
\begin{tabular}{@{}clSSSSScSSSS@{}}
	\toprule
	& {\multirow{2}{*}{Metric}} & {\multirow{2}{*}{Original}} & {\multirow{2}{*}{A4NT}}	& \multicolumn{3}{c}{{One-LM}}	 & \text{ } & \multicolumn{3}{c}{Intersection}	& {\multirow{2}{*}{Union}} \\
	\cmidrule{5-7} \cmidrule{9-11}  
	&                       &                            &                          & {No DB} & {DB = 25} & {DB = 40} && {No DB} & {DB = 25}  &  {DB = 40}  & \\
    \midrule
    \multirow{3}{*}{\STAB{Ling.}} & 
    BT Accuracy (\%)    & 100.00	&66.49	&94.47	&92.88	&90.60	&&95.41	&87.39	&88.63	&96.88 \\
    &GPT-2 PPL	        &41.71	&44.85	&39.51	&53.65	&66.21	&& 41.6	&42.80	&58.15	&42.07\\
    &Lex. Div. (\%)	    &3.22	&2.28	&2.52	&1.82	&1.09	&& 2.50	&1.47	&0.97	&2.71\\
    \midrule[0.1pt]
    \multirow{3}{*}{\STAB{Clsf.}} &
    Clsf. Accuracy (\%)	&64.73	&61.31	&62.07	&61.69	&59.52	&& 64.23	&60.90	&59.81	&64.02\\
    &Entropy	            &0.87	&0.86	&0.87	&0.91	&0.95	&&0.87	&0.93	&0.95	&0.87\\
    &CR (\%)	            &14.21	&15.72	&13.44	&6.49	&2.80	&&13.95	&4.78	&2.47	&14.22\\
	\bottomrule
\end{tabular}

%% file: tables/3dom.tex
  \newcolumntype{D}{>{\arraybackslash}m{0.18\linewidth}} 
\begin{tabular}{@{}clScSSScSSSS@{}}
	\toprule
	& {\multirow{2}{*}{Metric}} & {\multirow{2}{*}{Original}} & {\multirow{2}{*}{A4NT}}	& \multicolumn{3}{c}{{One-LM}}	 & \text{ } & \multicolumn{3}{c}{Intersection}	& {\multirow{2}{*}{Union}} \\
	\cmidrule{5-7} \cmidrule{9-11}  
	&                       &                            &                          & {No DB} & {DB = 25} & {DB = 40} && {No DB} & {DB = 25}  &  {DB = 40}  & \\
    \midrule
	\multirow{3}{*}{\STAB{Ling.}}
	&BT Accuracy (\%)	&100.00	& {--}    &93.84	&93.64	&87.83	&&89.09	&89.25	&82.47	&93.30  \\
    &GPT-2 PPL	        &41.70	& {--}    &43.49	&48.99	&84.61	&&48.15	&49.70	&69.08	&42.66 \\
    &Lex. Div. (\%)	    &3.41	& {--}    &2.46	    &1.81	&0.94	&&1.97	&1.02	&0.77	&2.86 \\
    \midrule[0.1pt]
    \multirow{3}{*}{\STAB{Clsf.}}
    &Clsf. Accuracy (\%)	&49.78	& {--}    &49.16	&47.64	&45.41	&&48.12	&47.13	&45.81	&48.81 \\
    &Entropy	            &1.38	& {--}    &1.38	    &1.43	&1.47	&&1.44	&1.44	&1.49	&1.38 \\
    &CR (\%)	           &43.00	& {--}    &43.33	&38.89	&30.75	&&38.76	&35.37	&28.02	&45.88 \\
	\bottomrule
\end{tabular}

%% file: body/related.tex
\vspace{-1ex}
\section{Related Work}
\vspace{-1ex}

A large body of prior work has attempted to address \textit{algorithmic bias} by modifying different stages of the natural language processing (NLP) pipeline.~\citet{blodgett-etal-2021-stereotyping},~\citet{barikeri-etal-2021-redditbias},~\citet{farrand2020neither},~\citet{mireshghallah-etal-2021-privacy} and~\citet{sheng2019woman} propose and analyze benchmarks for evaluating fairness in different  applications.
~\citet{ravfogel2020null},~\citet{kaneko2019gender},~\citet{shin2020neutralizing} and~\citet{kanekodebiasing} attempt to de-bias word embeddings used by NLP systems, while \citet{elazar2018adversarial, Barrett2019AdversarialRO,wang-etal-2021-dynamically} attempt to de-bias model representations and encodings. 

There is also a large body of work on modifying learning algorithms and inference procedures to produce more fair outcomes~\cite{agarwal2018reductions, madras2018predict, zafar2017fairness,han-etal-2021-decoupling,mireshghallah2021not}.  
While effective in many cases, such approaches do nothing to mitigate \textit{human bias} in decisions based on text. 
Fundamentally, our framework is concerned with stylistic features of human-generated text. Thus, a large body of prior work on methods for unsupervised style transfer are related to our approach~\cite{santos2018fighting,yang2018unsupervised,luo2019dual,he2020probabilistic}. There is also a vast body of work on style obfuscation~\cite{emmery2018style, reddy2016obfuscating, bevendorff2019heuristic, a4nt}. 

Our work is most closely related to~\citet{a4nt} and ~\citet{xu2019privacy}. A4NT~\cite{a4nt} attempts to obfuscate text style through automatic re-writing. However, their approach attempts to hide protected attributes simply by mapping the style of one protected category to that of another.
In contrast, we seek not to map the author's text to another author's style, but to a central obfuscated style. 
~\citeauthor{xu2019privacy} propose Privacy Aware Text Re-writing (PATR), which takes a similar adversarial learning translation based approach to address this problem and re-write text.
One fundamental difference between our style-pooling method and PATR is that we provide the choice of union vs. intersection of styles, which is concerned with the societal aspects of removing sensitive attributes, since we are targeting removal of bias. PATR, however, targets privacy and is therefore not concerned with the union vs. intersection of sensitive attributes.

Finally, there is a body of work on re-writing text to mitigate the potential biases within the content of the text itself.
~\citeauthor{Ma2020PowerTransformerUC} propose PowerTransformer, which rewrites text to correct the implicit and potentially undesirable bias in character portrayals. 
~\citeauthor{Pryzant2020AutomaticallyNS} propose a framework that addresses subjective bias in text and ~\citeauthor{field2020unsupervised} and ~\citeauthor{zhou2021challenges} introduce approaches to identifying gender bias against women at a comment level and dialect bias in text, respectively.
These works focus on the text content, and not on the stylistic features of the author.

%% file: body/conclusion.tex
\vspace{-1.5ex}
\section{Conclusion}
\vspace{-1.5ex}
We proposed a probabilistic VAE framework for automatically re-writing text in order to obfuscate stylistic features that might reveal sensitive attributes of the author. We demonstrated in experiments that our proposed framework can indeed reduce bias in downstream text classification. Finally, our model poses two ways of defining a central style. Future work might consider further explorations of alternative notions of stylistic centrality.

%


%% file: body/ackn.tex
\section*{Acknowledgments}
The authors would like to thank the anonymous reviewers and meta-reviewers for their helpful feedback. We also thank Junxian He for insightful discussions. Additionally, we thank our colleagues at the UCSD Berg Lab for their helpful comments and feedback. 

\section*{Ethical Considerations}

Our proposed model is intended to be used to address a real-world fairness issue. However, this is an extremely complicated topic, and it should be treated with caution, especially upon deploying possible mitigations such as ours. 
One potential issue we see is the chance that systems like this might obfuscate text by converging towards the majority and erasing styles of marginalized communities.
We have tried to address this concern, and raise discussion around it in our introduction and model design, by allowing for multiple operationalizations of a ``central'' style, and introducing the union and intersection obfuscations.
Defining a true notion of centrality that would  effectively protect sensitive attributes without erasing any specific styles of writing requires further study. 

%% file: body/appendix.tex
\section{Appendix}
\label{sec:appendix}


\subsection{Experiment Code}\label{sec:app-code}
We have uploaded code, data and model checkpoints needed for reproducing the experiments in \url{https://github.com/mireshghallah/style-pooling}. 
There, you will find a \texttt{Read Me} file which includes all the necessary steps and link to the data and model checkpoints. 
In short, you need to download the data and model checkpoints from the link. When you download the models-data compressed folder, extract it. Place the content of the data folder and the models in corresponding folders in the code.
The package dependencies are all included in the \texttt{dependencies} file. In order to create a similar setup, please install the exact version mentioned there.

Once all the directories are setup, you can train your own models using the commands in the \texttt{Read Me}, or you can evaluate the models we have already provided. 
Evaluation code is included in the \texttt{results\_ipynbs} folder. 

\subsection{Model Configurations}\label{sec:app-configs}

\paragraph{Seq2Seq Model.}

For all the experiments, We use single layer LSTMs with hidden size of 512 as both the encoder and decoder, and we use a word embedding size of 128.  We apply dropout to the readout states before softmax with a rate of 0.3. We add a max pooling operation over the encoder hidden states before feeding it to the decoder.

\paragraph{Language Model: Yelp data.} We use an LSTM language model with hidden size of 512 and word embedding size of 128 and dropout value of 0.3.

\paragraph{Language Model: Blog and Twitter data.} We use an LSTM language model with hidden size of 2048 and word embedding size of 1024 and dropout value of 0.3.

\paragraph{Sensitive-attribute Classifiers.} We use LSTM classifiers for classifying sensitive attributes. The hidden size is 512 and word embedding size is 128. The last layer size is the number of sensitive classes. 

\paragraph{GPT-2} We used this repository~\url{https://github.com/priya-dwivedi/Deep-Learning/tree/master/GPT2-HarryPotter-Training} to download and feed data to the GPT-2 model, and get the PPL score.

\subsection{Dataset Details}\label{sec:app-data}

\paragraph{Yelp.} The training set contains 399,999 sentences, and test set consists of 30,000 sentences, both divided equally between the 3 domains. The vocabulary size is 9.6k words. The misspelled words of $D_0$, $D_1$ and $D_2$ are ``00great00, 00this00, 00it00, 00to00, 00food00'', ``11of11, 11place11, 11for11, 11good11, 11service11'' and ``22they22, 22are22, 22in22, 22very22, 22my22'', respectively. 

\paragraph{Blogs.} The blogs dataset is a collection of micro blogs from \url{blogger.com}  which consists of 19,320 `documents' (over 3.3 million sentences) along with annotation of author’s age, gender, and occupation. Each document is a collection of an individual author's posts. We will use this data in both two and three domain style-pooling, where we treat age as the sensitive attribute and balance the data so each domain has the same number of sentences. In the two domain setup,  we divide the data in two groups of teenagers, $13-18$ and adults $23-48$. In the three style setup, we have three groups of teenagers  ($13-18$), young adults ($23-28$) and adults ($33-48$). The age groups $19-22$ and $29-32$ are missing from the data. After preprocessing and balancing the dataset, we end-up with 1.2 Milion sentences in the training set, 400k sentences in the test for the 2 domain setup, and 762k training sentences and 192k test sentences for the test set. There are 10k words in the vocabulary. All the datasets are balanced.

\paragraph{Twitter.} There are 146.5k sentences in the training set, and 11.2k sentences in the test set. We reproduced this data using this scripts from~\citet{elazar2018adversarial}'s GitHub repository:~\url{https://github.com/yanaiela/demog-text-removal}.

\subsection{Hyperparameters}\label{sec:app-hyperparams}
For all experiments, we set the training batch size to $32$, the test batch size to $128$ and the temperature of the softmax to $0.01$. 

\noindent{\textbf{KL weight hyperparameter:}}
The KL term in Eq.~\ref{eq:elbo} that appears naturally in the ELBO objective, can be treated as a regularizer that uses our $p_{prior}$ to induce the type of style we want. 
Therefore, in practice, we add a weight $\lambda$ to the KL term in ELBO since the regularization strength from our priors varies depending on the datasets, training data size, or prior structures\cite{bowman2016generating}.

\paragraph{Yelp.} For the Yelp experiments, the learning rate is set to $0.001$ and the KL weight ($\lambda$) for the Union, One-LM and Intersection experiments are $0.03$, $0.03$ and $0.02$, respectively. 

\paragraph{Blogs and Twitter.} For the Blogs experiments, the learning rate is $0.0005$, and the KL weight ($\lambda$) is $0.04$ (for both 2 and 3 domains).

\subsection{Comparison with A4NT Details}
To compare with the work ``A4NT: Author Attribute Anonymity by Adversarial Training of Neural Machine Translation''~\cite{a4nt}, we downloaded a checkpoint of their pre-trained model, available in their github repository: \url{https://github.com/rakshithShetty/A4NT-author-masking}. Since we have also used the same dataset with the same train/test separation, we use the model as is for evaluation. 

\subsection{Human Evaluation Experiment Details}

Our crowd workers are recruited from the Amazon Mechanical Turk (AMT) platform. Each HIT required the workers to answer a question regarding only one pair of sentences, and each worker was paid $\$0.1$ per HIT. For English proficiency, the workers were restricted to be from USA or UK. For the semantic consistency test, the question asked from the Turkers was: ``Which sentence is closer in meaning to the original sentence below?'', where the original sentence and the obfuscated ones where provided to the workers. For fluency, we asked: ``Which sentence is more fluent in English?''.

\subsection{Comparison with ~\citet{elazar2018adversarial}}\label{sec:app-elazar}

%
%
%

~\citet{elazar2018adversarial} aims at creating representations for text that could be used for a specific classification task, while hiding sensitive attributes.  
Although our approach deals with the text as opposed to representations and and can be applied for a wider range of downstream tasks, we   offer a brief comparison to this method.
~\citeauthor{elazar2018adversarial} use the Twitter dataset~\citet{rangel2016overview}, set the sensitive attribute to be age, and try to produce representations that would perform well on the main task of ``conversation detection'' (mention detection) on Tweets. 
On the original data, they report an accuracy of $77.5\%$ and $64.8\%$ for a classifier that tries to classify conversations and age, respectively, which drop to  $72.5\%$ and $57.3\%$, after applying their adversarial learning scheme.

We cloned their repository and used their code to process the dataset.We then created and trained the conversation and age classifiers, and reached an accuracy of $75.8\%$ and $64.63\%$ for them, respectively.  These dropped to  $73.28\%$ and $54.2\%$, after applying applying our intersection method.
This shows that for this particular task, our re-written text can out-perform prior work.


